\documentclass[10pt,twocolumn,letterpaper]{article}

\pdfoutput=1

\usepackage{cvpr}
\usepackage{times}
\usepackage{epsfig}
\usepackage{graphicx}
\usepackage{amsmath}
\usepackage{amssymb}

\usepackage[utf8]{inputenc} 
\usepackage[T1]{fontenc}    
\usepackage{booktabs}       
\usepackage{amsfonts}       
\usepackage{nicefrac}       
\usepackage{microtype}      

\usepackage[utf8]{inputenc}
\usepackage{tikz}
\usepackage{pgfplots}
\pgfplotsset{compat=1.3}

\usepackage{balance}

\usepackage{soul}
\usepackage{makeidx}
\usepackage{graphicx}
\usepackage{amsmath,amssymb} 
\usepackage{color}
\usepackage{subfig}
\usepackage{caption}
\usepackage{rotating}
\usepackage{array}
\usepackage{todonotes}
\usepackage{color, colortbl}
\usepackage{collcell}
\usepackage{hhline}
\usepackage{pgf}

\usepackage{enumitem}
\usepackage{algorithm}
\usepackage{algorithmic}

\newcommand {\rednote}[1]{}
\newcommand {\bluenote}[1]{}

\usepackage{mathtools}

\usepackage{amsthm}
\DeclarePairedDelimiter\abs{\lvert}{\rvert}%
\DeclarePairedDelimiter\norm{\lVert}{\rVert}%

\makeatletter
\let\oldabs\abs
\def\abs{\@ifstar{\oldabs}{\oldabs*}}
\let\oldnorm\norm
\def\norm{\@ifstar{\oldnorm}{\oldnorm*}}
\makeatother

\newcommand {\figref}[1]{Fig.~\ref{fig:#1}}

\newcommand {\tableref}[1]{Table.~\ref{table:#1}}

\newcommand {\sgn}{\text{sgn}}
\newcommand {\SD}{\emph{SD}}
\newcommand {\CE}{\emph{CE}}
\newcommand {\MSE}{\emph{MSE}}
\newcommand {\EMD}{\emph{EMD}}
\newcommand {\oEMD}{\overrightarrow{\emph{EMD}}}
\newcommand {\tEMD}{\emph{HEMD}}

\newcommand {\gEMD}{\nabla\emph{EMD}}

\cvprfinalcopy 

\def\cvprPaperID{1390} 

\ifcvprfinal\fi
\begin{document}

\title{Relaxed Earth Mover's Distances for Chain- and Tree-connected Spaces\\ and their use as a Loss Function in Deep Learning}


\author{Manuel Martinez$^{\ast}$, Monica Haurilet$^{\ast}$, Ziad Al-Halah$^{\ast}$, Makarand Tapaswi$^{\dag}$, Rainer Stiefelhagen$^{\ast}$\\[0.5em]
\begin{minipage}{\linewidth}\begin{center}\begin{tabular}{ccc}
$^{\ast}$Computer Vision for Human-Computer Interaction Lab && $^{\dag}$Machine Learning Group\\
Institute for Anthropomatics and Robotics && Department of Computer Science \\
Karlsruhe Institute of Technology && University of Toronto \\
{\tt\small cvhci.anthropomatik.kit.edu} && {\tt\small makarand@cs.toronto.edu } \\
\end{tabular}
\end{center}\end{minipage}
}

\maketitle


\begin{abstract}
The Earth Mover's Distance ($\EMD$) computes the optimal cost of transforming one distribution into another, given a known transport metric between them.
In deep learning, the $\EMD$ loss allows us to embed information during training about the output space structure like hierarchical or semantic relations.
This helps in achieving better output smoothness and generalization.
However $\EMD$ is computationally expensive.
Moreover, solving $\EMD$ optimization problems usually require complex techniques like \emph{lasso}.
These properties limit the applicability of $\EMD$-based approaches in large scale machine learning.

We address in this work the difficulties facing incorporation of $\EMD$-based loss in deep learning frameworks.
Additionally, we provide insight and novel solutions on how to integrate such loss function in training deep neural networks.
Specifically, we make three main contributions:
(i) we provide an in-depth analysis of the fastest state-of-the-art $\EMD$ algorithm (Sinkhorn Distance) and discuss its limitations in deep learning scenarios.
(ii) we derive fast and numerically stable closed-form solutions for the $\EMD$ gradient in output spaces with chain- and tree- connectivity; and
(iii) we propose a relaxed form of the $\EMD$ gradient with equivalent computational complexity but faster convergence rate.
We support our claims with experiments on real datasets.
In a restricted data setting on the ImageNet dataset, we train a model to classify 1000 categories using 50K images, and demonstrate that our relaxed $\EMD$ loss achieves better Top-1 accuracy than the cross entropy loss.
Overall, we show that our relaxed $\EMD$ loss criterion is a powerful asset for deep learning in the small data regime.
\end{abstract}



\begin{figure}[t]
\centering
\begin{tikzpicture}[scale=0.7]
\node[inner sep=-1pt] (giraffe) at (0,0) {\includegraphics[width=.05\textwidth]{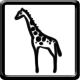}};
\node[inner sep=-1pt] (elephant) at (2,0) {\includegraphics[width=.05\textwidth]{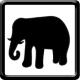}};
\node[inner sep=-1pt] (truck) at (4,0) {\includegraphics[width=.05\textwidth]{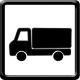}};
\node[inner sep=-1pt] (plane) at (6,0) {\includegraphics[width=.05\textwidth]{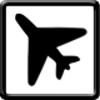}};
\node[inner sep=-.5pt, outer sep=-.5pt] (animal) at (1,2) {};
\node[inner sep=-.5pt, outer sep=-.5pt] (vehicle) at (5,2) {};
\draw[-,thick] (giraffe.north)     edge [bend left =25] node[anchor=center, above, sloped] {0.2 $\rightarrow$} (animal.south west);
\draw[-,thick] (elephant.north)    edge [bend right=25] node[anchor=center, above, sloped] {0.1 $\rightarrow$} (animal.south east);
\draw[-,thick] (truck.north)       edge [bend left =25] node[anchor=center, above, sloped] {$\leftarrow$ 0.3} (vehicle.south west);
\draw[-,thick] (plane.north)       edge [bend right=25] node[anchor=center, above, sloped] {$\leftarrow$ 0.2} (vehicle.south east);
\draw[-,thick] (animal.north east) edge [bend left =25] node[anchor=center, above, sloped] {0.1 $\rightarrow$} (vehicle.north west);

\node[inner sep=-1pt] (furgofante) at (-2.25,-1.25) {$f\left(\vcenter{\hbox{\includegraphics[width=.04\textwidth]{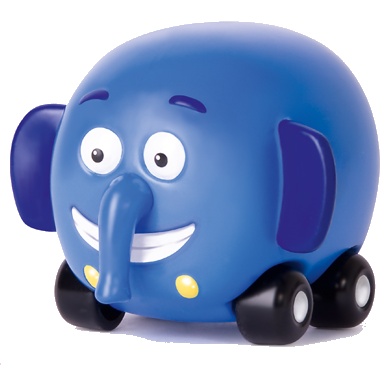}}}\right)$};
\node[inner sep=-1pt] (furgofantClass) at (-2.25,-2.25) {$y_{\vcenter{\hbox{\includegraphics[width=.02\textwidth]{icons/furgofante.jpg}}}}$};
\node[inner sep=-1pt] (emd1) at (-2.25,-3.25) {$\gEMD$};
\node[inner sep=-1pt] (emd2) at (-2.25,-4.25) {$\gEMD^2$};

\draw[-,thick] (-3,- .75) -- (7,- .75);
\draw[-] (-3,-1.75) -- (7,-1.75);
\draw[-,thick] (-3,-2.75) -- (7,-2.75);
\draw[-] (-3,-3.75) -- (7,-3.75);
\draw[-,thick] (-3,-4.75) -- (7,-4.75);

\draw[-,thick] (-1,  .25) -- (-1,-4.75);
\draw[-] (1,  .25) -- (1,-4.75);
\draw[-] (3,  .25) -- (3,-4.75);
\draw[-] (5,  .25) -- (5,-4.75);
\draw[-] (7,  .25) -- (7,-4.75);

\node at (0,-1.25) {  $0.2$}; \node at (2,-1.25) {  $0.4$}; \node at (4,-1.25) {  $0.2$}; \node at (6,-1.25) {  $0.2$};
\node at (0,-2.25) {  $0.0$}; \node at (2,-2.25) {  $0.5$}; \node at (4,-2.25) {  $0.5$}; \node at (6,-2.25) {  $0.0$};
\node at (0,-3.25) { $1.5$}; \node at (2,-3.25) { $-0.5$}; \node at (4,-3.25) {$-1.5$}; \node at (6,-3.25) {  $0.5$};
\node at (0,-4.25) {  $0.5$}; \node at (2,-4.25) { $-0.1$}; \node at (4,-4.25) { $-0.7$}; \node at (6,-4.25) {  $0.3$};

\end{tikzpicture}
\caption{
We train a deep learning model using \emph{Ellyvan} (a hybrid elephant and van).
The output of the forward pass is presented in the first row, and the label in the second.
The tree-structured output space allows us to train the network using gradients from the Earth Mover's Distance ($\gEMD$).
We introduce a regularized form that converges faster ($\gEMD^2$) and is more stable.
}
\label{fig:main}
\end{figure}

\section{Introduction}
The Wasserstein metric~\cite{villani2008optimal} is a distance function based on the optimal transport problem that compares two data distributions.
While computing such metrics on digital devices, it is common practice for data distributions to work in a discretized space (\eg~arranged in \emph{bins}).
Here, the Wasserstein distance is popularly known as the Earth Mover's Distance ($\EMD$)~\cite{rubner1998metric}.
The name is derived from a visual analogy of the data distributions as two piles of \emph{dirt} (earth).
$\EMD$ is defined as the minimum amount of effort required to make both distributions look alike.
Note that the individual bins of both distributions should be non-negative and their total mass equal (as is the case with probability distributions).

The $\EMD$ is widely used to compare histograms and probability distributions~\cite{marinai2011using,peleg1989unified,rolet2016fast,rubner1998metric,rubner2000earth}.
However, calculating the $\EMD$ is known to be computationally expensive.
This has led to several relaxed versions of the $\EMD$ for cases where speed is critical, \eg~when comparing feature vectors~\cite{ling2007efficient,pele2009fast,rabin2008circular}.

In addition to the large computational cost, $\EMD$ has the drawback of an $\ell_1$ behavior.
Solving $\EMD$ optimization problems often require \emph{lasso} optimization techniques (\eg, mirror descent, Bregman projections, etc.).
This represents a significant drawback for current deep learning approaches that strongly favor gradient-based methods such as Stochastic Gradient Descent, Momentum~\cite{sutskever2013momentum}, and Adam~\cite{ba2015adam}, that provide several small updates to the model parameters.


\vspace{0.1cm}
\noindent\textbf{Sinkhorn Distance.}
Using the $\EMD$ within an iterative optimization scheme was made feasible by Cuturi~\cite{cuturi2013sinkhorn}, who realized that an entropically regularized $\EMD$ can be efficiently calculated using the Sinkhorn-Knopp~\cite{sinkhorn1967diagonal} algorithm.
The resulting is referred to as the Sinkhorn Distance ($\SD$), and has achieved wide popularity within a number of learning frameworks~\cite{benamou2015iterative,bonneel2015sliced,cuturi2016smoothed,frogner2015learning,montavon2015wasserstein,rabin2015convex,rolet2016fast,solomon2015convolutional,solomon2014earth,solomon2014wasserstein}.
The $\SD$ approximates $\EMD$ effectively, and provides a subgradient for the $\EMD$ as a side result of the estimation.
This $\SD$ subgradient has been used to train deep learning models~\cite{frogner2015learning} and is implemented as a loss criterion in popular deep frameworks such as Caffe~\cite{jia2014caffe} and Mocha~\cite{Mocha}.
However, as $\SD$ is an $\ell_1$ norm, Frogner~\etal~\cite{frogner2015learning} need to combine $\SD$ with the Küllback-Leibler divergence and use an exceedingly small learning rate for it to converge.

Furthermore, the $\SD$ algorithm is prone to numerical instabilities when used in deep learning frameworks.
In some conditions, these instabilities imply that $\SD$ is not a close approximation of $\EMD$.
We believe that an analysis of the causes of instabilities is critical to extend the use of $\SD$ to deep learning frameworks and discuss them in detail in Sec.~\ref{sinkhorn}.

\vspace{0.1cm}
\noindent\textbf{Earth Mover's Distance.}
Concurrently, we suggest an alternative approach to $\SD$.
Instead of tackling the general case, we focus on output spaces whose connectivity graph takes the form of a chain (histograms or probability distributions) or a tree (hierarchies).
We provide closed-form solutions for the real $\EMD$ and its gradient. 


We start with chain-connected distributions (see \figref{chain}) that have a well-known closed-form solution~\cite{vallender1974calculation}, and derive its gradient.
We also propose a relaxed version of the $\EMD$, named $\EMD^2$ that exhibits similar structure but converges faster due to its $\ell_2$ behavior.


Furthermore, we derive a closed form solution for the $\EMD$ and its gradient that is valid for all metric spaces that have a tree connectivity graph.
This allows us to represent complex output spaces that are hierarchical in nature (\eg, WordNet~\cite{miller1995wordnet} and ImageNet~\cite{russakovsky2015ilsvrc}, Sentence Parse Trees~\cite{marcus1993penntreebank}).
We see an example of a hierarchical output space of object categories in \figref{main}.
We depict the expected flow of \emph{dirt} on the tree branches, and present the gradients for both original and relaxed versions of the $\EMD$ (details of the gradients in Sec.~\ref{subsec:relax_emd} and Sec.~\ref{treesection}).

\vspace{0.1cm}
\noindent\textbf{$\EMD$ as a loss criterion for deep learning.}
Using $\EMD$ as a loss criterion has several unique advantages over unstructured losses.
It allows us to shape the output relationships we expect from a model.
For example, it can tell the model that confusing a cat for a tiger may more acceptable than confusing a cat for a starship, and thus adds knowledge to the model.
Additionally, $\EMD$ gradients (in contrast to $\MSE$) are holistic and affect the whole output space as it is connected.
Therefore, models that predict the entire output space (\eg, histograms) converge faster.

Overall, we see that $\EMD$ has the effect of magnifying the information that an input data sample provides.
Each input sample does not only provide information about its own class, but also contains the relationship it has with the rest of the output bins (\eg, classes, histogram bins, etc.)
This second source of information helps generalize better, and results in improved performance with less data.

We demonstrate these characteristics in two real-world experiments.
We train a model to predict the Power Spectral Density of respiratory signals of sleep laboratory patients.
In this setting the $\EMD$ converges faster than the $\SD$ and $\MSE$ losses, and achieves better accuracy.
Our second experiment is performed on a reduced version of the ILSVRC 2012 challenge~\cite{russakovsky2015ilsvrc}.
We use all 1000 categories, but limit the training data to 50K images.
This, along with the $\EMD$ criterion, forces the network to learn the output space hierarchy.
While $\EMD$ alone is not enough to achieve the best top-1 accuracy, an equal combination of $\EMD$ and cross entropy loss achieves better top-1 accuracy than using cross entropy alone.








\begin{figure}[t]
  \centering
  \includegraphics[width=6cm]{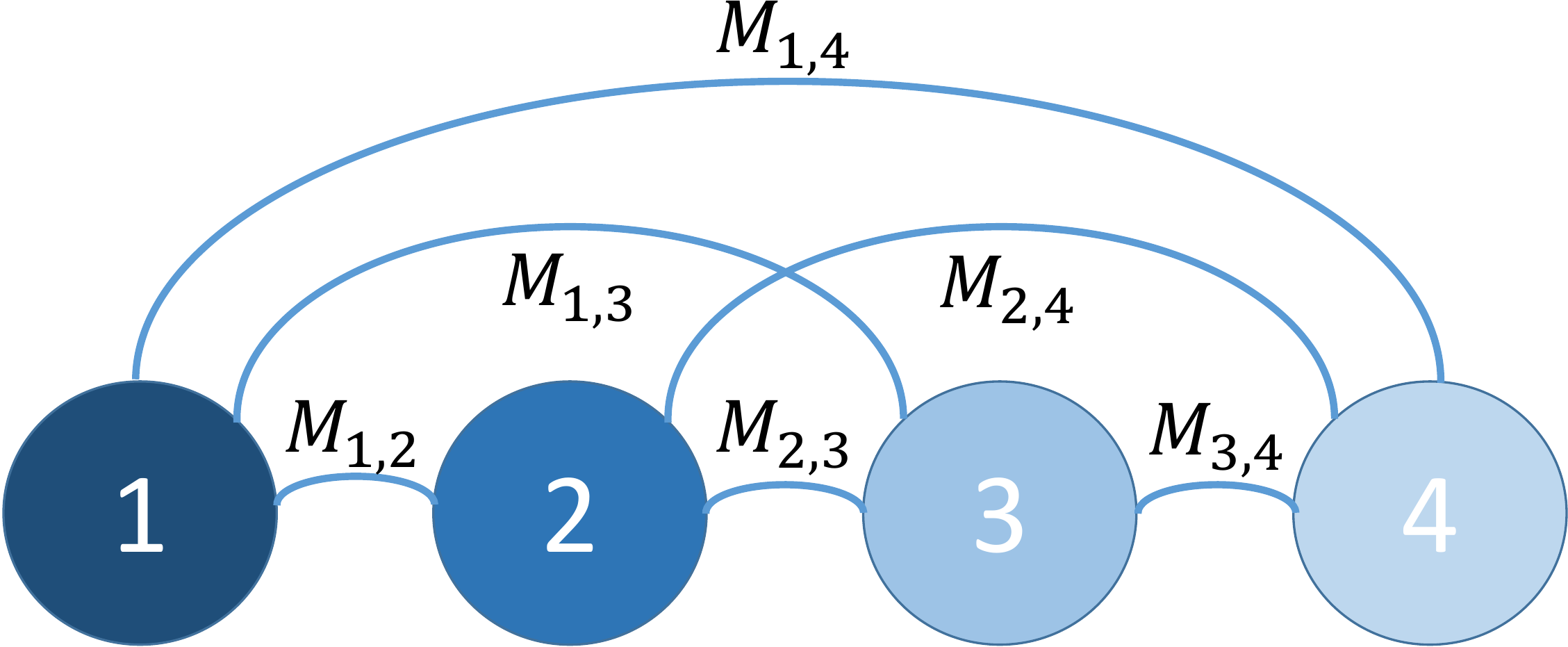}
  \caption{
  We illustrate the general case of the $\EMD$, where the \emph{dirt} can be moved from and to all bins directly.
  Computing the amount of \emph{dirt} that goes (flows) through each path is computationally expensive.}
  \label{fig:general}
\end{figure}


\begin{figure*}[t]
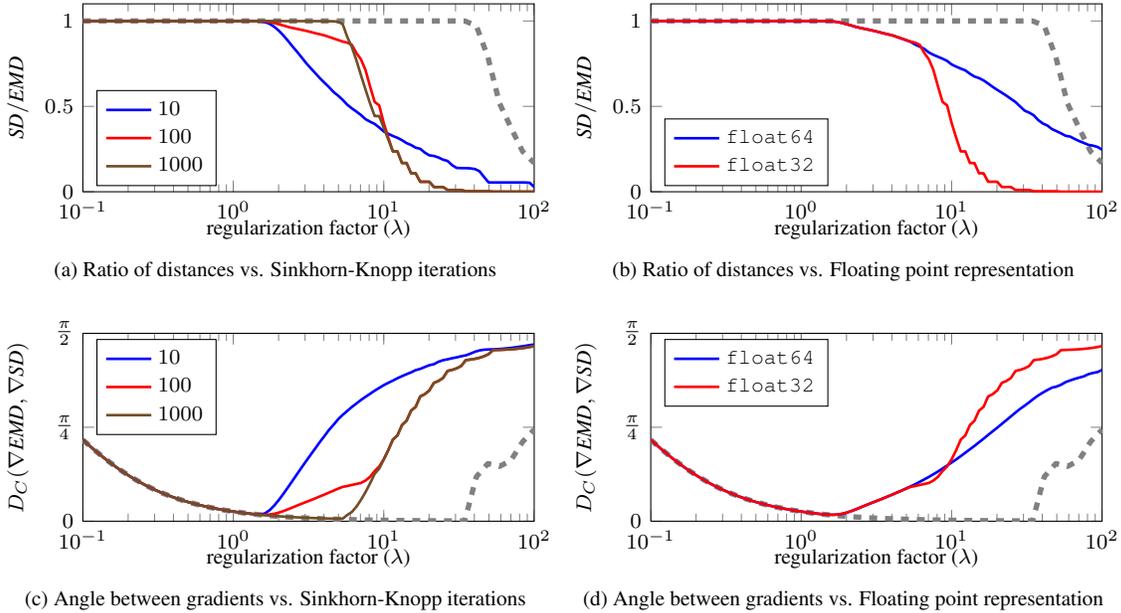

\centering

\subfloat[Ratio of distances vs. Sinkhorn-Knopp iterations]{
\begin{tikzpicture}
\tikzstyle{every node}=[font=\footnotesize]
\begin{semilogxaxis}[   height=2.5cm,  width=6cm,
scale only axis, ymin=0,ymax=1.1,xmin=.1,xmax=100 ,enlargelimits=false, y label style={at={(axis description cs:-0.1,.5)},anchor=south},  ylabel=\SD$ / $\EMD,  x label style={at={(axis description cs:0.5,-0.1)},anchor=north},  xlabel=regularization factor ($\lambda$), legend cell align=left, legend style={legend pos=south west,font=\scriptsize}]
\input{results/sinkhornFI}
\legend{$10$,$100$,$1000$}
\end{semilogxaxis}
\end{tikzpicture}}
\subfloat[Ratio of distances vs. Floating point representation]{
\begin{tikzpicture}
\tikzstyle{every node}=[font=\footnotesize]
\begin{semilogxaxis}[   height=2.5cm,  width=6cm,
scale only axis, ymin=0,ymax=1.1,xmin=.1,xmax=100 ,enlargelimits=false, y label style={at={(axis description cs:-0.1,.5)},anchor=south},  ylabel=\SD$ / $\EMD,  x label style={at={(axis description cs:0.5,-0.1)},anchor=north},  xlabel=regularization factor ($\lambda$), legend cell align=left, legend style={legend pos=south west,font=\scriptsize}]
\input{results/sinkhornFT}
\legend{\texttt{float64},\texttt{float32}}
\end{semilogxaxis}
\end{tikzpicture}}

\subfloat[Angle between gradients vs. Sinkhorn-Knopp iterations]{
\begin{tikzpicture}
\tikzstyle{every node}=[font=\footnotesize]
\begin{semilogxaxis}[   height=2.5cm,  width=6cm,
scale only axis, ymin=0,ymax=90,xmin=.1,xmax=100 ,enlargelimits=false, ytick={0,45,90}, yticklabels={$0$,$\frac{\pi}{4}$,$\frac{\pi}{2}$}, y label style={at={(axis description cs:-0.1,.5)},anchor=south},  ylabel={$D_C(\nabla\EMD,\nabla\SD)$},  x label style={at={(axis description cs:0.5,-0.1)},anchor=north},  xlabel=regularization factor ($\lambda$), legend cell align=left, legend style={legend pos=north west,font=\scriptsize}]
\input{results/sinkhornGI}
\legend{$10$,$100$,$1000$}
\end{semilogxaxis}
\end{tikzpicture}}
\subfloat[Angle between gradients vs. Floating point representation]{
\begin{tikzpicture}
\tikzstyle{every node}=[font=\footnotesize]
\begin{semilogxaxis}[   height=2.5cm,  width=6cm,
scale only axis, ymin=0,ymax=90,xmin=.1,xmax=100 ,enlargelimits=false, ytick={0,45,90}, yticklabels={$0$,$\frac{\pi}{4}$,$\frac{\pi}{2}$}, y label style={at={(axis description cs:-0.1,.5)},anchor=south},  ylabel={$D_C(\nabla\EMD,\nabla\SD)$},  x label style={at={(axis description cs:0.5,-0.1)},anchor=north},  xlabel=regularization factor ($\lambda$), legend cell align=left, legend style={legend pos=north west,font=\scriptsize}]
\input{results/sinkhornGT}
\legend{\texttt{float64},\texttt{float32}}
\end{semilogxaxis}
\end{tikzpicture}}

\caption{We show how implementation and hyper-parameter choices affect $\SD$ with an experiment on the WordNet hierarchy for 1000 ImageNet classes.
(a,b): Ratio of $\SD$ to the real $\EMD$.
(c,d): Angle of the cosine distance $\left(D_C(\cdot,\cdot)\right)$ between $\EMD$ and $\SD$ gradients.
(a,c): Impact of limiting Sinkhorn-Knopp iterations. We see that larger values of $\lambda$ require a large amount of iterations to converge (two typical values used in deep learning are 10~\cite{frogner2015learning} and 100~\cite{jia2014caffe}).
(b,d): \texttt{float32} (GPU) has a reduced dynamic range compared to \texttt{float64} (CPU), which degenerates for large $\lambda$ values.
The gray dashed line shown as a reference, corresponds to $10000$ iterations using \texttt{float64} representation.
The most commonly cited value of $\lambda$ is 10.
}
\label{fig:sinkhorn}
\end{figure*}

\section{The Earth Mover's Distance}
\label{sinkhorn}

As discussed earlier, the $\EMD$ is defined for discrete distributions.
Here, the probability mass (or \emph{dirt}) is distributed in discrete piles or bins.
The effort of moving a mound of \emph{dirt} between two bins is a non-negative cost which is linearly proportional to the amount of \emph{dirt} and distance between the bins.

Within this discrete domain, the general form of $\EMD$ between two distributions $\mathbf{p}, \mathbf{q} \in \mathbb R^N_{+}$ with $\norm{\mathbf{p}}_1 = \norm{\mathbf{q}}_1$ is
\begin{equation}
\label{eq:emd_general}
\EMD(\mathbf{p}, \mathbf{q}) = \inf_{T \in U(\mathbf{p}, \mathbf{q})} \langle M, T \rangle \, ,
\end{equation}
where $\langle \cdot, \cdot \rangle$ is the Frobenius inner product and $M \in \mathbb R^{N \times N}_{+}$ defines the generalized distance between bins (see \figref{general}).
$U(\mathbf{p}, \mathbf{q})$ is the set of valid transport plans between $\mathbf{p}$ and $\mathbf{q}$,
\begin{equation}
\label{eq:transport_domain}
U(\mathbf{p}, \mathbf{q}) = \{ T \in \mathbb R^{N \times N}_{+} : T \cdot \mathbf{1}_N = \mathbf{p},\, T^\top \cdot \mathbf{1}_N = \mathbf{q} \} \, .
\end{equation}
$\mathbf{1}_N$ is an $N$ dimensional vector of all ones, and $T$ is constrained such that its row sum corresponds to the distribution $\mathbf{p}$ and column sum to $\mathbf{q}$.

Without loss of generality (a simple scalar normalization), in the rest of the paper we assume $\norm{\mathbf{p}}_1 = \norm{\mathbf{q}}_1 = 1$.

\subsection{Unnormalized distributions}
The original $\EMD$ is not defined for $\norm{\mathbf{p}}_1 \ne \norm{\mathbf{q}}_1$.
Although there are several ways to modify the $\EMD$ for unnormalized distributions~\cite{chizat2015unbalanced,frogner2015learning,pele2009fast} we consider that this goes against the spirit of the metric.
Therefore, prior to computing the $\EMD$, we $\ell_1$-normalize the input distributions either using a $\ell_1$ normalization layer, or a \emph{softmax} layer.

\subsection{Sinkhorn distance}

The general formulation of the $\EMD$ (Eq.~\ref{eq:emd_general}) is solved using linear programming which is computationally expensive.
However, this problem was greatly alleviated by Cuturi~\cite{cuturi2013sinkhorn} who suggested a smoothing term for the $\EMD$ in the form of
\begin{equation}
\label{eq:sd}
\SD_\lambda(\mathbf{p}, \mathbf{q}) = \inf_{T \in U(\mathbf{p}, \mathbf{q})} \langle M, T \rangle - \frac{1}{\lambda}\langle T, \log T \rangle \, ,
\end{equation}
which allows to use the Sinkhorn-Knopp algorithm~\cite{sinkhorn1967diagonal} to obtain an iterative and efficient solution.

The Sinkhorn-Knopp algorithm is notable as it converges fast and produces a subgradient for the $\SD$ without extra cost.
This subgradient can in turn be used to update parameters of machine learning models~\cite{frogner2015learning}.
The algorithm is defined as\footnote{Of the several variants of the algorithm, this follows the one in Caffe~\cite{jia2014caffe}. $\odot$ and $\oslash$ stand for per-element multiplication and division respectively.}:

\begin{algorithm}                      
\caption{Sinkhorn Distance and Gradient}          
\label{alg1}                           
\begin{algorithmic}                    
    \REQUIRE $f(x), y, \lambda, \mathbf{M}$
    \STATE $\mathbf{K} \leftarrow e^{-\lambda \mathbf{M}-1}$, $u \leftarrow \mathbf{1}$,  $it \leftarrow 1$
    \WHILE{not converged \AND it $\le$ MAX-ITER}
		\STATE $u \leftarrow f(x) \oslash \left( \mathbf{K} ( y \oslash \mathbf{K}^{\top}  u) \right)$
    \ENDWHILE
    \STATE $v \leftarrow y \oslash (\mathbf{K}^{\top}  u)$
    \STATE $\SD \leftarrow \text{sum}( u \odot (\mathbf{K} \odot \mathbf{M} v))$
    \STATE $\nabla\SD \leftarrow \log{u}/\lambda$
\end{algorithmic}
\end{algorithm}

\subsection{Numerical stability of the Sinkhorn Distance}

We claim that the $\SD$ is not numerically stable when used in common deep learning frameworks.
We substantiate this claim by comparing the output of the $\SD$ and its gradient to the real $\EMD$.
We extract a hierarchy of categories from the WordNet ontology for the 1000 classes of the ILSVRC2012 dataset~\cite{russakovsky2015ilsvrc}.
The tree has 1374 nodes in total.
This hierarchy acts as the structure of the output space in our evaluation.

There are three parameters that impact numerical stability of the $\SD$ in deep learning:
\begin{description}[noitemsep,nolistsep]
\item [Iteration limit:] any practical implementation needs an upper limit for the number of $\SD$ iterations.
This leads to a trade-off between speed and accuracy which is more apparent for larger values of $\lambda$, as seen in \figref{sinkhorn} (a), (c).
\item [Floating point accuracy:] the Sinkhorn algorithm alternately normalizes rows and columns of the transport matrix.
This requires several multiply-accumulate operations which are prone to numerical inaccuracies.
This problem is made worse by the exponential form of $\mathbf{K} = e ^{-\lambda \mathbf{M} -1}$, which increases the dynamic range of the values.
To make things worse, GPUs use a \texttt{float32} representation, instead of the common \texttt{float64} representation used by CPUs.
In \figref{sinkhorn} (b) and (d), we observe how using \texttt{float32} affects the results, especially for large values of $\lambda$ where more iterations are required to converge.
\item [Regularization factor ($\lambda$):] the regularization factor affects both the accuracy and the convergence behavior of $\SD$.
In a non-deep learning framework, the number of iterations does not pose a limit and \texttt{float64} representation can be used.
Lower values of $\lambda$ imply better convergence behavior~\cite{cuturi2013fast}, while larger values approximate the Earth Mover's Distance better (see gray reference line in \figref{sinkhorn}).
However, in deep learning applications where \texttt{float32} representation are common and the iteration number is typically chosen to be $10$~\cite{frogner2015learning} or $100$~\cite{jia2014caffe}, larger values of $\lambda$ are unusable.
\end{description}

In general, $\SD$ works best when the ratio between the largest and the smallest non-zero value in the $\mathbf{M}$ is small (thus has a smaller dynamic range), and the size of the output space is small (due to the reduced number of multiply-add operations).


\begin{figure}[t]
  \centering
  \subfloat[Chain $\EMD$]{\includegraphics[width=6cm]{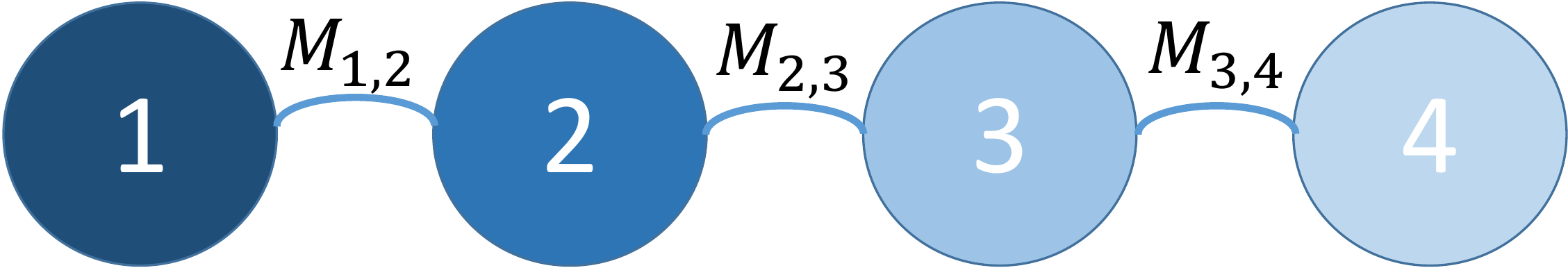}}
  \caption{
  One dimensional probability distributions are a perfect example of chain output spaces.
  In chains, \emph{dirt} must travel through all intermediate bins in its path.
  This greatly simplifies the computation of the $\EMD$, as it only requires to eliminate one node and edge at each step.
  }
  \label{fig:chain}
\end{figure}

\section{$\EMD$ in chain-connected spaces}
\label{oneDimLoss}

We analyze the scenario where bins in a distributions (\eg~histograms, probabilities) are situated in a one dimensional space.
Here, moving \emph{dirt} from a source to target bin requires an ordered visit to every bin in between (see \figref{chain}).

The bin distance $M$ can be defined recursively to ensure that only consecutive bin distances are considered.
\begin{equation}
M_{i,j} =
\begin{cases}
0 &\text{if } i = j \, ,\\
M_{i-1,j} + d_{i-1,i} &\text{if } i > j \, ,\\
M_{j,i} &\text{if } i < j \, .
\end{cases}
\end{equation}
Here, $d_{i-1, i}$ is the distance between two consecutive bins (typically all equal and $1$).

The above choice of bin distances facilitates a simple solution to calculate the $\EMD$ using a recursion.
Essentially, each bin either receives all the excess \emph{dirt} that results from leveling previous bins, or in case of a deficit, tries to provide for it.
Note that the cost of going left-to-right ($M_{i,j}$) or right-to-left ($M_{j,i}$) is symmetric.
The closed form recursive formulation for the $\EMD$ between two one-dimensional distributions is:
\begin{equation}
\label{eq:emd}
\oEMD(\mathbf{p}, \mathbf{q}) = \sum_{i=1}^{N-1} M_{i,i+1} \cdot | \varphi_i | \, ,
\end{equation}
where $\varphi_i$ represents the excess \emph{dirt} that needs to be moved ahead or deficit in \emph{dirt} that needs to be filled up to bin $i$.
\begin{equation}
\varphi_i =
    \sum_{j=1}^i \left(
		  p_j - q_j
    \right) \, .
\end{equation}
For notational brevity, we will refer to $M_{i,i+1}$ as $\hat{M}_i$.
The above expression can be rewritten with the sign function as
\begin{equation}
\oEMD(\mathbf{p}, \mathbf{q}) =  \sum_{i=1}^{N-1} \hat{M}_i  \cdot \sgn(\varphi_i) \cdot \varphi_i \, .
\end{equation}

Note that as both distributions have the same amount of total mass, and we progressively level out the \emph{dirt} over all bins, when we arrive to the last bin all \emph{dirt} will have been leveled (\ie~$\varphi_N = 0$).
Therefore, we compute the outer sum only up to $N-1$.

\subsection{Gradient of the Earth Mover's Distance}

To integrate $\EMD$ as a loss function in an iterative gradient-based optimization approach, we need to compute the analytical form of the gradient.
However, we must ensure that the gradient obeys the law of \emph{dirt} conservation.
The gradient should not create new or destroy existing \emph{dirt} to avoid changing the total mass of the distributions ($\norm{\mathbf{p}}_1 \neq 1$ after updates).

We use the trick of projected gradients, and define $\mathbf{e}_k$ as a vector of length $N$ whose value at entry $k$ is $1-1/N$, and $-1/N$ elsewhere.
Note that $\mathbf{e}_k$ sums to $0$.
For a small value $h$, we compute the distance between the perturbed distribution $\mathbf{p} + h\mathbf{e}_k$ and $\mathbf{q}$ as:

\begin{equation}
\begin{split}
 & \oEMD(\mathbf{p} + h\mathbf{e}_k, \mathbf{q}) \simeq   \\
 &  \sum_{i=1}^{N-1} \hat{M}_i \cdot \sgn(\varphi_i)
    \sum_{j=1}^i \left( p_j + h(\delta_{jk} - 1/N) - q_j \right) \, ,
\end{split}
\end{equation}
where $\delta_{jk} = 1$ iff $j = k$.
Note that by choosing $h$ small enough, $\sgn(\varphi_i)$ can be assumed to remain unchanged.
The corresponding partial derivative for the $\oEMD$ is:
\begin{eqnarray}
\label{eq:gradZeroSum_emd}
\frac{\partial \oEMD( {\mathbf{p}}, {\mathbf{q}} )} {\partial p_k} \simeq
\label{eq:gemd_l1}
  \sum_{i=1}^{N-1} \hat{M}_i \cdot \sgn (\varphi_i)  \sum_{j=1}^i \left(\delta_{jk} - 1/N \right) \, .
\end{eqnarray}





\subsection{Relaxed Earth Mover's Distance}
\label{subsec:relax_emd}

The proposed gradient (Eq.~\ref{eq:gemd_l1}) is numerically stable and avoids erosion or addition of new dirt.
This is an important step forward to use $\EMD$ in learning frameworks.

However, as the distance contains an absolute value function ($|\varphi_i|$) we observe difficulties converging to a solution, similar to $\ell_1$ optimization.
In particular, when $\hat{M}_i \in \mathbb{N}$, it is easy to see that the terms of $\gEMD$ are integer fractions of $N$ (see \figref{main}, multiples of 0.25).
Furthermore, as small changes to the distribution do not change the gradient, the Hessian is zero except at a few discrete set of points (where the sign of $\varphi_i$ changes) making optimization hard.

To solve these issues, we suggest a relaxed form of the $\EMD$ where the cost is calculated proportional to a power of the excess/deficit of \emph{dirt}.
\begin{equation}
\label{eq:emd_rho}
\oEMD^\rho(\mathbf{p}, \mathbf{q}) = \sum_{i=1}^{N-1} \hat{M}_i \cdot | \varphi_i |^\rho \, .
\end{equation}
whose gradient is:
\begin{eqnarray}
\label{eq:g_emd2_l1}
\nabla\oEMD^\rho \simeq  \rho \sum_{i=1}^{N-1} \hat{M}_i  \cdot \varphi_i \cdot | \varphi_i |^{\rho-2} 
  \sum_{j=1}^i \left( {\delta_{jk} - 1/N} \right) \, .
\end{eqnarray}

For $\rho=1$ we have the normal $\EMD$ distance, which behaves like $\ell_1$.
During gradient descend we suggest to use the case with $\rho=2$ that bears similarity with the popular Mean Squared Error loss.
$\EMD^2$ preserves the nice properties of conserving dirt, while having real valued gradients (see an example in \figref{main}).
In addition, we see that $\oEMD^2$ also exhibits non-zero Hessians. 

\begin{equation}
\label{eq:hessian_emd2_l1}
\begin{split}
 \frac{\partial^2 \oEMD^2( {\mathbf{p}}, {\mathbf{q}} )} {\partial p_k \partial p_l} =  2 \sum_{i=1}^{N-1} \hat{M}_i & \cdot (N \cdot H(i-l) - i) \\
 & \cdot (N \cdot H(i-k) - i) \, ,
\end{split}
\end{equation}

where $H: \mathbb{R} \rightarrow \{0, 1\}$ is the Heaviside step function defined as $H(n) = \{1 \text{ if } n \geq 0, 0 \text{ otherwise}\}.$


\subsection{Discussion: comparing $\EMD$, $\SD$ and $\MSE$}
Plotting the gradients provides a good impression of the actual behavior of the $\EMD$.
It also shows how $\EMD$ serves as a criterion that provides holistic optimization over the output space (full distribution).

\begin{figure*}[ht!]
	\centering
	\subfloat[Mean Squared Error]{\includegraphics[width=5.5cm]{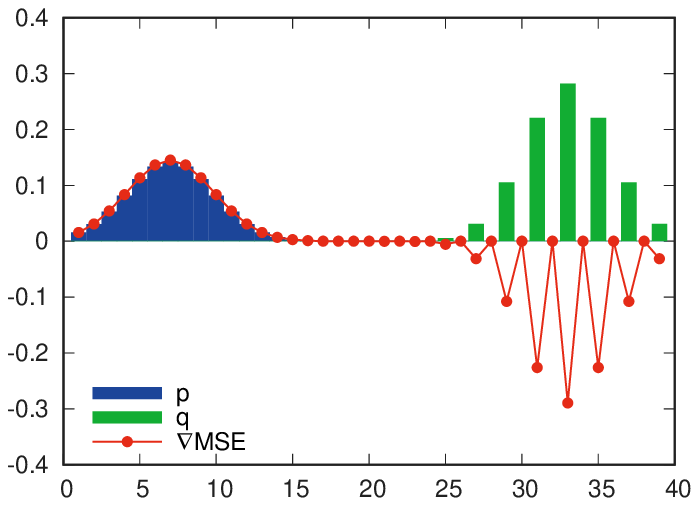}}\quad
	\subfloat[Earth Mover's Distance]{\includegraphics[width=5.5cm]{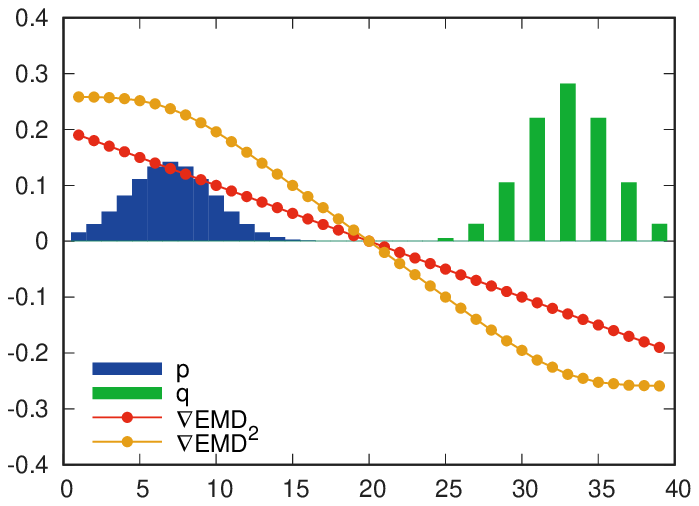}}\quad
	\subfloat[Sinkhorn Distance]{\includegraphics[width=5.5cm]{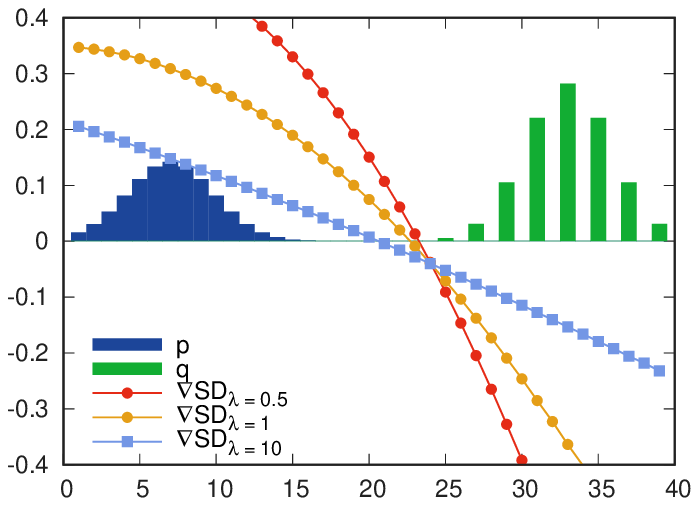}}
	\caption{Gradient flow to convert a source distribution $\mathbf{p}$ to a target $\mathbf{q}$.
    (a) $\MSE$ gradient is not smooth, and does not affect the whole output space.
    (b) $\EMD$ and $\EMD^2$ gradients affect the whole space. $\EMD^2$ has $\ell_2$ behavior, while $\EMD$ behaves like $\ell_1$.
    (c) $\SD$ gradients also affect the whole space but the regularized versions are asymmetric. $\SD_{\lambda = 10}$ approximates $\EMD$ well.}
	\label{fig:resultsB}
\end{figure*}

In \figref{resultsB} we show the gradients corresponding to several different loss criteria for the transformation between two unit-norm distributions: a smooth one $\mathbf{p}$, and a spiky one $\mathbf{q}$.
We present the gradient of $\MSE$ in \figref{resultsB} (a).
Note how $\MSE$ optimizes each bin independently, and results in a non-smooth gradient.

In \figref{resultsB} (b) we show the gradients for $\EMD$ and $\EMD^2$.
In both cases the gradient is holistic and affects the whole output space.
Furthermore, the regularization effect induced by $\EMD^2$ results in a smoother gradient.

In \figref{resultsB} (c) we show the $\SD$ gradient for $\lambda$ values of 0.5, 1 and 10.
In all cases the gradients are also holistic.
We see that that larger values of $\lambda$ produce a gradient resembling that of the true $\EMD$, however it comes with its own set of problems that were discussed earlier.


\section{$\EMD$ in tree-connected spaces}
\label{treesection}

We demonstrate here how $\EMD$ can be used to model output spaces with a tree structure.
Our formulation expects that all observed bins correspond to the leaves of the tree, and the remaining latent nodes and have no \emph{dirt}.
As we can link a tree to any non-leaf node with a zero-cost connection, this formulation allows us to express any tree structure.
We refer to this analysis as the Hierarchical Earth Mover's Distance ($\tEMD$) (see \figref{main}).
Note that, this is still compatible with all our previous developments (as chains are a sub-class of trees).
As such, we do not distinguish between $\EMD$ and $\tEMD$ while presenting the evaluation.

We define $\tEMD^\rho$ as:
\begin{equation}
\label{eq:smd}
\tEMD^\rho(\mathbf{p}, \mathbf{q}) = \sum_{i \in G} M_{i,\mathbf{p}(i)} \cdot | \tilde{\varphi}_i |^\rho,
\text{ where }
\end{equation}
\begin{equation}
\tilde{\varphi}_i =
    \sum_{j \in \mathbf{l}(i)} \left(
		  p_j -  q_j
    \right) \, ,
\end{equation}
where $M_{i,\mathbf{p}(i)}$ is the cost of transporting \emph{dirt} from node $i$ to its parent (abbreviated as $\tilde{M}_i$),
$G$ is the  set of all nodes in the tree,
and $\mathbf{l}(i)$ is the set of all leaves in the subtree that has $i$ as a root.
$N$ is the total number of leaves (and bins) in the tree.
The intuition behind this formula is that we can reduce the tree one leave at a time as if it were the tail of a chain.

Then the gradient of $\tEMD$ is defined as:

\begin{eqnarray}
\label{eq:g_emd2}
\nabla\tEMD^\rho \simeq  \rho \sum_{i \in G} \tilde{M}_i \cdot \tilde{\varphi}_i \cdot | \tilde{\varphi}_i |^{\rho-2}
  \sum_{j=1}^i \left( {\delta_{jk} - 1/N} \right) \, .
\end{eqnarray}

All equations can be solved efficiently by a post-order traversal of nodes in the tree.




\section{Experimental analysis}

We implement our models, the $\EMD$ and the $\SD$ criterions using Torch~\cite{collobert2011torch7}%
\footnote{Code will be made publicly available}.
Evaluation is performed on an i5-6600K CPU at 3.5GHz with 64GB of DDR4-2133 RAM and a GTX1080 GPU running Ubuntu 16.04, CUDA 8.0 and cuDNN 5.1.3.
Unless otherwise stated, the $\SD$ hyper-parameters are $\lambda=3$, the iteration limit $100$, and using CUDA (\ie~\texttt{float32} type).

\subsection{Timing analysis for $\SD$ vs. $\EMD^2$}
\begin{table}[t]
\centering
\begin{tabular}{ r | c | c | c |}
     & max iter. & CPU   & GPU \\ \hline
	$\SD$ &   10 & 942ms & 15.1ms\\
	$\SD$ &  100 & 7.51s & 88.9ms\\
	$\SD$ & 1000 & 74.5s & 865ms\\
	$\EMD^2$ & - & 126ms & 25ms\\ \hline
\end{tabular}
\caption{Computation time for the gradients of $\SD$ and $\EMD^2$ on the WordNet Tree Structure experiment for one minibatch of size 512.
The closed form solution allows the $\EMD^2$ to be 60x faster to calculate than 100 iterations of $\SD$ on a CPU.
Our unoptimized CUDA code is 3.5x faster than $\SD$.
Note that it is not practical to run $\SD$ on \texttt{float64} precision and 1000 iterations, as it takes 74 seconds to evaluate the loss for a single minibatch.}
\label{table:speed}
\end{table}

We evaluate the computational efficiency of $\EMD^2$ and $\SD$.
The Sinkhorn-Knopp algorithm is very efficient and demonstrates fast GPU performance.
However, being an iterative procedure, $\SD$ is significantly slower than $\EMD^2$ (see \tableref{speed}).
Furthermore, it is not practical for large output spaces, especially if we require the use of \texttt{float64} precision which is only available on CPUs.

\subsection{$\EMD^2$ on Chain Spaces}

We evaluate the use of $\EMD$ to learn Power Spectral Density (PSD) - a chain distribution with signal power binned into different frequencies.
As we need to optimize over the whole output space, this task is not only well-suited to use $\EMD$ as a loss criterion, but $\EMD$ also serves as the evaluation metric.

Our task is to predict the PSD of a breathing signal obtained from a patient using chest excursion signals (a nose thermistor acts as reference).
Our data is recorded from 75 real patients from a sleep laboratory.
We extract 200 one-minute clips for each patient, providing us with a dataset of 15,000 samples.
We use data from 60 patients to train our models, and the remaining 15 as test subjects.

Noise levels depend on the activity of the patient, and are negligible when he/she is relaxed.
On the other hand, when the patient moves, sits, or talks during the 1 minute segment, the correlation between chest movement and respiration disappears.

We adopt a two layer network for this experiment.
The first layer consists of $16$ temporal convolution filters with a receptive field of $11$.
We apply the \emph{tanh} nonlinearity, and stack a fully connected layer on top.
To ensure positive outputs (as we predict signal power) we apply the \emph{square} function $(\cdot)^2$ to the output layer.
We use the \emph{Adam} optimizer~\cite{ba2015adam}.

\begin{figure}[t]
\centering
\resizebox{\columnwidth}{!}{%
\begin{tikzpicture}
\tikzstyle{every node}=[font=\footnotesize]
\begin{axis}[   height=3cm,  width=8cm, title=Learning to predict the PSD of a breathing signal,
x filter/.code={\pgfmathparse{#1/60}\pgfmathresult},
scale only axis, ymin=0,ymax=10,xmin=0,xmax=12 ,enlargelimits=false, y label style={at={(axis description cs:-0.05,.5)},anchor=south},  ylabel=Test EMD,  x label style={at={(axis description cs:0.5,-0.1)},anchor=north},  xlabel=Epoch, legend cell align=left, legend style={legend pos=north east,font=\scriptsize}]
\input{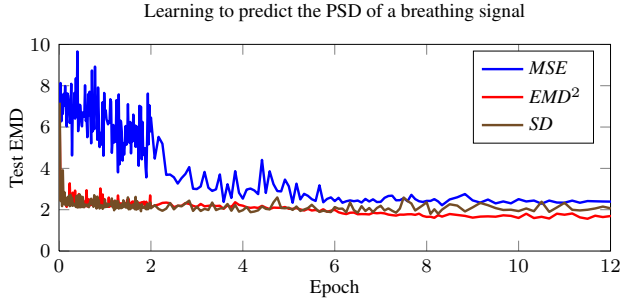}
\legend{$\MSE$,$\EMD^2$,$\SD$}
\end{axis}
\end{tikzpicture}}
\caption{We train a regressor to estimate the PSD of real breathing signals obtained from chest excursions.
We observe that both $\EMD^2$ and $\SD$ learn the transformation significantly faster than the $\MSE$.
Over a longer period $\EMD^2$ achieves better accuracy than $\SD$.
}
\label{fig:resultsPSD}
\end{figure}

On this simple task \emph{Adam} performs very well, and the model converges with all criteria (see \figref{resultsPSD}).
Nevertheless, both $\EMD^2$ and $\SD$ outperform $\MSE$, converging in a fraction of the first epoch.

This highlights the benefits of using $\EMD$ criterion in cases where it can be hard to obtain several training samples and the output space has a suitable structure.

\begin{figure*}[t]
\centering
\begin{tikzpicture}
    \begin{axis}[%
    hide axis,xmin=10,xmax=50,ymin=0,ymax=0.4,
	legend columns=4,legend style={cells={align=left}}
    ]
    \addlegendimage{line width=1.5pt, purple}
    \addlegendentry{$\emph{CE} \quad$};
    \addlegendimage{line width=1.5pt, blue}
    \addlegendentry{$\EMD^2 \quad$};
    \addlegendimage{line width=1.5pt, teal}
    \addlegendentry{$50\% \enskip \EMD^2 \quad$\\$50\% \enskip \emph{CE}$};
    \addlegendimage{line width=1.5pt, olive}
    \addlegendentry{$25\% \enskip \SD \quad$\\$75\% \enskip \emph{CE}$};
    \end{axis}
\end{tikzpicture}\\
\vspace{-4.5cm}
\subfloat[Top-1 Accuracy on ImageNet@1280K images]{
\begin{tikzpicture}
\tikzstyle{every node}=[font=\footnotesize]
\begin{axis}[   height=2.6cm,  width=4.55cm,
y filter/.code={\pgfmathparse{100*(1-#1)}\pgfmathresult},
scale only axis, ymin=0,ymax=50,xmin=2,xmax=100 ,enlargelimits=false, y label style={at={(axis description cs:-0.075,.5)},anchor=south},  ylabel=Top-1 Accuracy,  x label style={at={(axis description cs:0.5,-0.1)},anchor=north},  xlabel=Epoch, legend cell align=left, legend style={at={(0.5,1.1)},anchor=south,font=\tiny}]
\addplot+[purple, line width=1pt, mark=none, line join=round] table[x expr=\coordindex+2, y index=1] {results/FULL_CE_ErrorRate1.log};
\addplot+[blue,   line width=1pt, mark=none, line join=round] table[x expr=\coordindex+2, y index=1] {results/FULL_EMD_ErrorRate1.log};
\addplot+[teal,   line width=1pt, mark=none, line join=round] table[x expr=\coordindex+2, y index=1] {results/FULL_EMD2CE_ErrorRate1.log};
\addplot+[olive,  line width=1pt, mark=none, line join=round] table[x expr=\coordindex+2, y index=1] {results/FULL_SD4_ErrorRate1.log};
\end{axis}
\end{tikzpicture}}
\subfloat[Top-1 Accuracy on ImageNet@50K images]{
\begin{tikzpicture}
\tikzstyle{every node}=[font=\footnotesize]
\begin{axis}[   height=2.6cm,  width=4.55cm,
y filter/.code={\pgfmathparse{100*(1-#1)}\pgfmathresult},
scale only axis, ymin=0,ymax=10,xmin=2,xmax=200 ,enlargelimits=false, y label style={at={(axis description cs:-0.075,.5)},anchor=south},  ylabel=Top-1 Accuracy,  x label style={at={(axis description cs:0.5,-0.1)},anchor=north},  xlabel=Epoch, legend cell align=left, legend style={at={(0.5,1.1)},anchor=south,font=\tiny}]
\addplot+[purple, line width=1pt, mark=none, line join=round] table[x expr=\coordindex+2, y index=1] {results/CE_0_LR005_ErrorRate1.log};
\addplot+[blue,   line width=1pt, mark=none, line join=round] table[x expr=\coordindex+2, y index=1] {results/EMD_0_ErrorRate1.log};
\addplot+[teal,   line width=1pt, mark=none, line join=round] table[x expr=\coordindex+2, y index=1] {results/EMD2_0_ErrorRate1.log};
\addplot+[olive,  line width=1pt, mark=none, line join=round] table[x expr=\coordindex+2, y index=1] {results/SD4_0_ErrorRate1.log};
\end{axis}
\end{tikzpicture}}
\subfloat[Holistic Error on ImageNet@50K images]{
\begin{tikzpicture}
\tikzstyle{every node}=[font=\footnotesize]
\begin{axis}[   height=2.6cm,  width=4.55cm,
scale only axis, ymin=0,ymax=10,xmin=2,xmax=200 ,enlargelimits=false, y label style={at={(axis description cs:-0.075,.5)},anchor=south},  ylabel=$\EMD$,  x label style={at={(axis description cs:0.5,-0.1)},anchor=north},  xlabel=Epoch, legend cell align=left, legend style={at={(0.5,1.1)},anchor=south,font=\tiny}]
\addplot+[purple, line width=1pt, mark=none, line join=round] table[x expr=\coordindex+2, y index=0] {results/CE_0_LR005_Wasserstein.log};
\addplot+[blue,   line width=1pt, mark=none, line join=round] table[x expr=\coordindex+2, y index=0] {results/EMD_0_Wasserstein.log};
\addplot+[teal,   line width=1pt, mark=none, line join=round] table[x expr=\coordindex+2, y index=0] {results/EMD2_0_Wasserstein.log};
\addplot+[olive,  line width=1pt, mark=none, line join=round] table[x expr=\coordindex+2, y index=0] {results/SD4_0_Wasserstein.log};
\end{axis}
\end{tikzpicture}}
\caption{We analyze the use of Earth Mover's Distance as a loss criterion on the 1000-class ImageNet Large-Scale Visual Recognition Challenge.
(a) We evaluate on the full training set, where a combined loss of $\EMD^2$ and Cross Entropy provides the best top-1 accuracy.
$\EMD^2$ alone does not achieve high top-1 accuracy as it tries to optimize for the whole output space and not only the best result.
(b) We evaluate on 50K images, which is about $4\%$ of the original training set.
In this case the improvement provided by $\EMD^2$ criterion is more apparent.
(c) We plot the holistic error of the entire output space for the 50K images subset.
The $\EMD^2$ outperforms others and its combinations by a large gap.
}
\label{fig:imagenet}
\end{figure*}
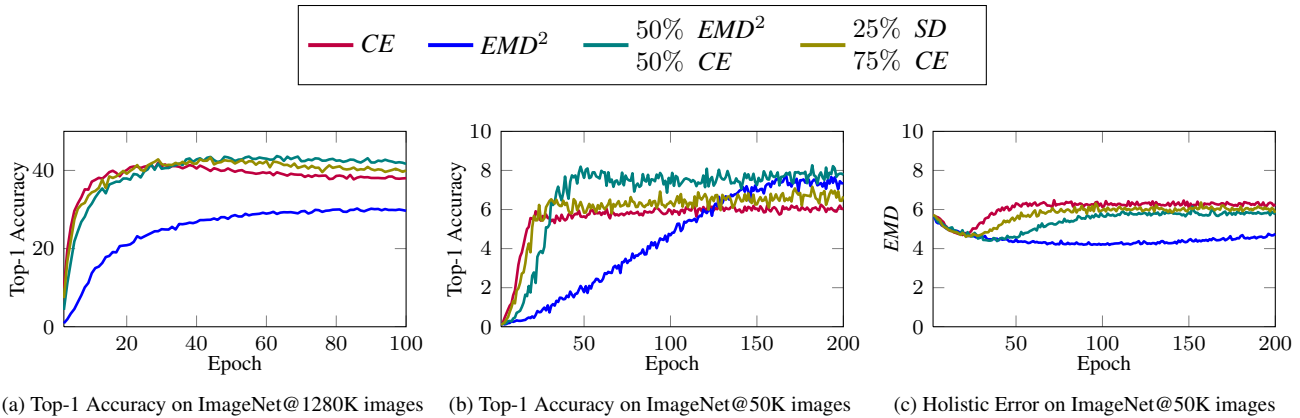


\subsection{$\EMD^2$ on Tree Spaces}

To evaluate the $\EMD^2$ loss on a hierarchical space, we develop an experiment based on the well known 1000-class ImageNet object recognition challenge~\cite{russakovsky2015ilsvrc}.

We train a model similar to Alexnet~\cite{krizhvesky2012alexnet} with batch normalization~\cite{ioffe2015batch} after ReLU, using a minibatch size of $512$, a learing rate of $0.05$ with a decay of $10^{-5}$ and a $\ell_2$ weight penalty of $10^{-4}$.
We use Stochastic Gradient Descent (SGD) as optimizer with momentum of $0.9$.
The input image is downsized to $112 \times 112$ pixels, and horizontal flipping and cropping is used for data augmentation at train time only.

The output space hierarchy tree is obtained from WordNet~\cite{miller1995wordnet} and has a total of $1374$ nodes and a maximum distance between nodes of $26$. We set all edge costs to $1$.
By our definition, the output labels correspond to the leaves of the tree.
Thus, the minimum hierarchical distance between a pair of output labels is $2$.

We evaluate the following loss criteria:
\begin{description}[noitemsep,nolistsep]
\item [Cross Entropy ($\CE$):] the standard loss used in classification problems and state-of-the-art ImageNet models.
\item [$\EMD^2$:] pure $\EMD^2$ after a softmax non-linearity.
\item [$\EMD^2 + \CE$:] a $1:1$ combination of $\EMD^2$ and $\CE$.
\item [$\SD + \CE$:] a $1:3$ combination of $\SD$ and $\CE$. We give more emphasis to $\CE$ since using a $1:1$ ratio did not converge.
\end{description}

$\SD$ and $\EMD$ alone do not converge using SGD on a large variation of parameter combinations that we tried.
This is somewhat expected behavior for $\ell_1$ losses.

We explore two data setups.
In the first case, the full training set is available (1280K images), while in the second, only a small amount of training data is available ($\sim4\%$, 50K images).
Our results show little advantage for $\EMD$ when using the full training set (see \figref{imagenet}~(a)).
This happens because the model can already learn the output space hierarchy from the input images used for training.
However, obtaining such large datasets is a daunting task.

We discuss the results at more depth for the second setting with reduced data (50K images).
As metrics, we present the Top-1 accuracy and the $\EMD$ loss.
Top-1 accuracy is a common metric used in the ImageNet challenge which depends only on the largest value of the output vector.
On the other hand, the $\EMD$ loss has an opposite notion, as it depends on the output of the entire vector.

\noindent\textbf{$\CE$ loss:}
The $\CE$ loss strongly favors Top-1 accuracy, and thus converges the fastest with regards to the Top-1 metric (see \figref{imagenet}~(a) and (b)).
When operating in the reduced data setting, improvements in Top-1 accuracy have the side effect of reducing the holistic loss (\figref{imagenet}~(c)) as the model also begins to learn the leaves of the hierarchical space from the input data.
However, the model soon starts over-fitting, the Top-1 accuracy plateaus, and the holistic loss grows back to it's original value.

\noindent\textbf{$\EMD^2$ loss:}
We see an effect opposite to that of $\CE$ when using $\EMD^2$.
The $\EMD^2$ loss optimizes primarily for the holistic loss (see \figref{imagenet}~(c)).
Here, the improvements in Top-1 accuracy are a side effect and also slow.
Nevertheless, we see in \figref{imagenet}~(b) that the Top-1 accuracy of $\EMD^2$ ends up higher than $\CE$ in the 50K image setting, as it learns the entire output space.

\noindent\textbf{$\EMD^2$ + $\CE$ losses:}
The combination of losses provides fast network convergence and highest Top-1 accuracy (see \figref{imagenet}~(b))
The $\CE$ loss optimizes Top-1 accuracy while the $\EMD^2$ incorporates the information of the output space through the holistic optimization.

\noindent\textbf{$\SD$ + $\CE$ losses:}
We see to a lesser extent a similar behavior to that of $\EMD^2$ + $\CE$.
However, the performance is limited by the fact that $\SD$ is mainly an $\ell_1$ distance, and the Sinkhorn-Knopp algorithm is not numerically stable for large output spaces with the \texttt{float32} representation.

\section{Conclusion}

$\SD$ has achieved wide-spread popularity in applications that wish to optimize for the $\EMD$ criterion.
However, with a thorough analysis, we point out two limitations in $\SD$ which make it hard to use in deep learning:
(i) numerical instability due to floating point precision; and
(ii) the $\ell_1$ behavior that makes it hard to optimize.

We counter this by deriving closed-form solutions for $\EMD$ and its \emph{dirt} conserving gradient on chain (\eg~histograms) and tree (\eg~hierarchies) output spaces.
We also propose a relaxed version ($\EMD^2$) of the original distance and compute its analytical form.
Our $\EMD^2$ exhibits better properties regarding numerical stability and convergence.

On a task about predicting the PSD of respiratory signals (chain-connectivity), we demonstrate faster convergence and reduction in error using $\EMD^2$.
We also evaluate object categorization on 1000 classes from the ImageNet challenge and work in the regime of limited training data (50K image samples).
Here, using the WordNet hierarchy (tree-connectivity), we observe that modeling the output space through the use of $\EMD^2$ helps boost the performance.

Our contributions will help promote a wider adaption $\EMD$ as a loss criterion within deep learning frameworks.

\balance

{\small
\bibliographystyle{ieee}
\bibliography{commonbib/longstrings,commonbib/common}
}

\end{document}